\documentclass[letterpaper, 10 pt, conference]{ieeeconf}  

\IEEEoverridecommandlockouts                              

\RequirePackage{xcolor, hyperref} 
\usepackage{cite}
\usepackage[normalem]{ulem}

\usepackage{graphicx}
\usepackage{epsfig} 
\usepackage{textcomp}
\usepackage{pifont}

\usepackage{amsmath,amssymb,amsfonts}
\usepackage{algorithm}
\usepackage{algpseudocode}
\usepackage{multirow}
\usepackage{booktabs}
\usepackage[flushleft]{threeparttable}

\usepackage{xcolor}
\usepackage{colortbl}
\usepackage{hyperref}
\hypersetup{
  colorlinks=true,         
  linkcolor=blue, 
  citecolor=blue,
  filecolor=magenta,
  urlcolor=black  
}

\newcommand{\cc}[1]{\cellcolor[HTML]{EFEFEF}}
\newcommand{\rc}[1]{\rowcolor[HTML]{EFEFEF}}
\algrenewcommand\algorithmicrequire{\textbf{Input:}}
\algrenewcommand\algorithmicensure{\textbf{Output:}}
\algrenewcommand\algorithmiccomment[1]{// {\itshape #1}}

\definecolor{subsectioncolor}{rgb}{0.0, 0.5, 0.5} 

\newcommand{\customref}[3][]{%
  \text{#2}~%
  \ifx\\#1\
    \hyperref[#3]{\ref*{#3}}%
  \else%
    \hyperref[#3]{\ref*{#3}(#1)}%
  \fi
}

\useunder{\uline}{\ul}{}

\title{\LARGE \bf
Uncertainty-Aware Cross-Modal Knowledge Distillation with Prototype Learning for Multimodal Brain-Computer Interfaces
}

\author{
Hyo-Jeong Jang$^{1}$, Hye-Bin Shin$^{1}$, and Seong-Whan Lee$^{2}$
\thanks{*This research was supported by the Institute of Information \& Communications Technology Planning \& Evaluation (IITP) grant, funded by the Korea government (MSIT) (No. RS-2019-II190079 (Artificial Intelligence Graduate School Program (Korea University)), and No. RS-2024-00457882 (AI Research Hub Project)).}%
\thanks{$^{1}$H.-J. Jang and H.-B. Shin are with the Department of Brain and Cognitive Engineering, Korea University, Anam-dong, Seongbuk-ku, Seoul 02841, Korea.
{\tt\small \{h\_j\_jang, hb\_shin\}@korea.ac.kr}.}%
\thanks{$^{2}$S.-W. Lee is with the Department of Artificial Intelligence, Korea University, Anam-dong, Seongbuk-ku, Seoul 02841, Korea. 
{\tt\small sw.lee@korea.ac.kr}.}%
}

\begin{document}

\maketitle
\thispagestyle{empty}
\pagestyle{empty}


\begin{abstract}
Electroencephalography (EEG) is a fundamental modality for cognitive state monitoring in brain-computer interfaces (BCIs). However, it is highly susceptible to intrinsic signal errors and human-induced labeling errors, which lead to label noise and ultimately degrade model performance. To enhance EEG learning, multimodal knowledge distillation (KD) has been explored to transfer knowledge from visual models with rich representations to EEG-based models. Nevertheless, KD faces two key challenges: modality gap and soft label misalignment. The former arises from the heterogeneous nature of EEG and visual feature spaces, while the latter stems from label inconsistencies that create discrepancies between ground truth labels and distillation targets. This paper addresses semantic uncertainty caused by ambiguous features and weakly defined labels. We propose a novel cross-modal knowledge distillation framework that mitigates both modality and label inconsistencies. It aligns feature semantics through a prototype-based similarity module and introduces a task-specific distillation head to resolve label-induced inconsistency in supervision. Experimental results demonstrate that our approach improves EEG-based emotion regression and classification performance, outperforming both unimodal and multimodal baselines on a public multimodal dataset. These findings highlight the potential of our framework for BCI applications.
\end{abstract}

\section{INTRODUCTION}
Electroencephalography (EEG) is one of the neural monitoring modalities widely used in brain–computer interfaces (BCIs)~\cite{review_paradigms, eeg2020}.
As EEG reflects the brain's electrical activity in real time, it serves as one of the most direct physiological indicators of a user's cognitive and emotional states~\cite{aBCI3, aBCI}.   
However, EEG signals are inherently noisy and uncertain, owing to factors such as sensor artifacts, subject variability, experimental limitations, and the absence of a clearly defined ground truth~\cite{noise2, neurograsp, leeCurling}. 
These factors collectively give rise to semantic ambiguity into EEG interpretation, which is especially problematic in emotion recognition due to the prevalence of label noise~\cite{aBCI2, emotion2015}.

To complement unimodal EEG-based systems, multimodal BCIs have emerged as a promising approach to enhance decoding accuracy by integrating diverse signals~\cite{Li2025MultimodalBCI}.
Recent studies have proposed algorithmic strategies to incorporate additional modalities to enhance EEG signal interpretation~\cite{MultiModal1}. 
Various approaches have been explored to exploit these complementary signals, including knowledge distillation (KD), multimodal feature fusion, contrastive learning, adversarial learning, and co-training~\cite{caf, v2e, ekd}. 
Such multimodal learning techniques aim to improve robustness and generalization by integrating different modalities that capture the same underlying user state or intention.
Among them, KD~\cite{kd-hinton, kd-survey} has emerged as a powerful technique for transferring structured knowledge from a high-quality teacher modality to a lower-quality student modality~\cite{image1996}. 
This transfer allows the student model to overcome deficiencies in its input modality and improve performance.

Despite the promise of KD, it faces two critical challenges in multimodal learning: modality gap and soft label misalignment. The modality gap~\cite{CMKD2023ICLR} arises from the heterogeneous nature of representations across modalities. Specifically, while conventional knowledge distillation methods have shown effectiveness in unimodal settings, they often fail to generalize across modalities. Soft label misalignment~\cite{crosskd2024} occurs when noisy labels introduce uncertainty, leading to inconsistencies between the ground truth and the distillation target provided by the teacher model. When learning from both, the student may receive conflicting signals and either down-weight or disregard the teacher’s predictions~\cite{KDD2024}.

In the present study, we propose a cross-modal KD framework that leverages the visual modality to guide EEG feature learning through semantically informed representation alignment. Our framework incorporates two key components.
First, we employ a prototype-based similarity module that bridges the modality gap by projecting EEG and visual features into a shared latent space, reducing representational discrepancies and enabling alignment based on semantic confidence~\cite{prototype2024}. 
Second, our cross-modal KD component addresses soft label misalignment by interpreting the student’s intermediate features through the teacher’s head, enabling the transfer of informative signals that would be disregarded due to prediction-label inconsistencies~\cite{CMKD2022, crosskd2024}.

We evaluate our framework on a public multimodal emotion dataset, demonstrating its superiority over state-of-the-art unimodal and multimodal baselines in both discrete and continuous emotion recognition tasks.
Through ablation studies, we analyze the contribution of each loss component to the overall performance enhancement of our proposed framework.
By capturing information overlooked by conventional KD, our method improves robustness and semantic reliability in EEG-based emotion recognition.

\begin{figure*}[!t]
\centerline{\includegraphics[scale=0.8]{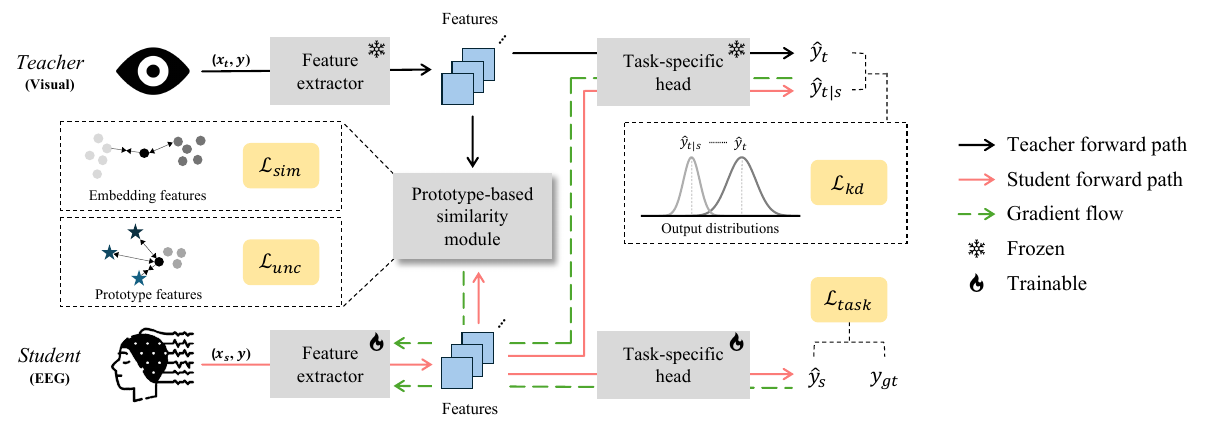}}
\caption{Overview of the proposed cross-modal KD framework. Intermediate features are extracted from both the teacher and student networks via their respective feature extractors. These features are passed through a prototype-based similarity module, which facilitates evidence-based uncertainty estimation and class-level semantic alignment. Additionally, the intermediate feature from the student is injected into the pre-trained teacher's task-specific distillation head to obtain predictions $\hat{y}_{t|s}$, which are then compared with the teacher's own predictions $\hat{y}_t$.}
\label{fig1}
\end{figure*}

\section{PROPOSED METHOD}

\subsection{Problem Definition}
In this paper, we address the problem of learning robust EEG representations under label noise and data uncertainty by distilling and transferring knowledge gained from the visual modality. The dataset contains paired EEG and visual samples $\{x_s^i, x_t^i, y^i\}_{i=1}^N$. Here, $x_s$ and $x_t$ represent data from the student (EEG) and teacher (visual) modalities, respectively, and $y$ denotes the corresponding ground truth labels. The feature extractors $\mathcal{E}$ and $\mathcal{V}$ map raw EEG and visual signals into their respective feature spaces. Also, EEG label noise lead to discrepancies between the ground truth labels and the distillation target from the teacher model, which negatively affects the student model's training.
Our goal is to learn a function $f_\theta : \mathcal{X}_s \rightarrow \mathcal{Y}$ that maps EEG representations to labels while leveraging visual knowledge. 
This is achieved by minimizing the total loss $\mathcal{L}$, which integrates all loss components in our framework.

\subsection{Proposed Framework}
This work aims to mitigate modality discrepancy and label noise in EEG-based learning. To begin with, we introduce the prototype-based similarity module, which builds upon prior work~\cite{prototype2024} using prototype-based similarity to improve cross-modal feature alignment.
In our framework, this module performs prototype-based similarity alignment guided by Dirichlet-based uncertainty estimation, allowing the model to align modality-specific representations while explicitly accounting for semantic ambiguity. Given EEG samples $x_s$ and visual samples $x_t$, we extract their features $f_s$, $f_t$ and corresponding embeddings $\mathbf{e}_s$, $\mathbf{e}_t$ as follows:
\begin{equation}\label{eq:feat}
    f_s,\mathbf{e}_s = \mathcal{E}(x_s), \ f_t,\mathbf{e}_t = \mathcal{V}(x_t), 
\end{equation} 
where $\mathcal{E}$ and $\mathcal{V}$ denote the EEG and visual feature extractors, respectively. 
We construct a similarity matrix $Q$ using cosine similarity to measure the pairwise relationships between EEG and visual embeddings, which serves as the basis for cross-modal feature alignment and is computed as:
\begin{equation}\label{eq:cosim}
Q_{ii}(\mathbf{e}_s, \mathbf{e}_t)=\beta \cdot \left({\frac{\mathbf{e}_s^i}{||\mathbf{e}_s^i||}}\right) \cdot \left({\mathbf{e}_t^i \over ||\mathbf{e}_t^i||}\right)^T.
\end{equation}
Here, $Q\in\mathbb{R}^{N \times N}$ denotes the pairwise similarities between EEG and visual samples in a batch, and $\beta$ is a temperature parameter. The diagonal elements $Q_{ii}$ correspond to matching EEG-visual pairs, while the off-diagonal elements $Q_{ij}$ indicate non-matching pairs. To align EEG representations with visual features, we define the similarity-based loss function using a softmax-based contrastive (InfoNCE) formulation:
\begin{equation}\label{eq:loss_sim}
\mathcal{L}_{sim} = - {1 \over N} \sum_{i=1}^N \log {{\exp(Q_{ii}(\mathbf{e}_s, \mathbf{e}_t))} \over {\sum_{k=1}^N \exp(Q_{ik}(\mathbf{e}_s, \mathbf{e}_t))}}.
\end{equation}
By optimizing $\mathcal{L}_{sim}$,  the model is encouraged to align EEG and visual features by maximizing the similarity between matched pairs.  
However, this loss does not account for the semantic ambiguity arising from label noise and misalignment issues. To address this limitation, we incorporate Dirichlet-based uncertainty estimation, which quantifies the model's confidence in each sample by measuring its similarity to class prototypes $\Phi$. 
Specifically, the concentration parameter of the Dirichlet distribution is computed as $\alpha_i = \exp(Q(\mathbf{e}, \Phi) / \tau) + 1$ for each class $i$, where the similarity matrix is defined as:
\begin{equation}\label{eq:cosim2}
Q_{ij}(\mathbf{e}, \Phi) = \beta \cdot \left({\mathbf{e}^i \over ||\mathbf{e}^i||}\right) \cdot \left({\Phi^j \over ||\Phi^j||}\right)^T.
\end{equation}
This formulation estimates sample reliability via the alignment between embeddings and class prototypes $\Phi$, where the total evidence is computed as $e_i = \sum_{j=1}^c \alpha_{ij}$ with $c$ denoting the number of prototypes. The uncertainty is calculated from this evidence, with $\tau$ as a temperature parameter:
\begin{equation}\label{eq:unc}
u_i= 1-{c \over e_i}= 1-{c \over {\sum_{j=1}^c (\exp({Q_{ij} \over \tau})+1)}}. 
\end{equation}
As the total evidence increases, the corresponding aleatoric uncertainty decreases, indicating higher confidence in the prediction. Thus, $u$ reflects the inverse of total evidence and serves as a principled measure of aleatoric uncertainty.
To integrate this uncertainty into training, we define the uncertainty-aware loss that aligns model-inferred uncertainty with observed similarity-based uncertainty. In this loss, $h_j$ denotes the average similarity of sample $j$ to all non-matching samples, weighted by a scaling factor $\delta$:
\begin{equation}\label{eq:loss_unc}
\mathcal{L}_{unc} = {1 \over N} \sum_{j=1}^N || u_j- (\delta h_j)||^2_2.
\end{equation}

Subsequently, we employ a task-specific distillation head. Inspired by~\cite{crosskd2024}, which reinterprets student features through the teacher’s structure in a unimodal setting, we extend this idea to a cross-modal scenario, where the teacher and student operate in different modalities. To minimize the semantic discrepancy caused by the modality gap, we feed the student’s intermediate features into the teacher’s head, specifically at the layer just before the final projection. This allows the student to benefit from the teacher’s role in guiding semantic alignment, even when the teacher’s predictions do not fully correspond to the ground truth. Accordingly, the outputs of the teacher and student heads are defined as follows:
\begin{equation}\label{eq:logits}
\hat{y}_t = \mathcal{H}_t(f_t), \ \hat{y}_s = \mathcal{H}_s(f_s), 
\end{equation}
where $\mathcal{H}_t$ and $\mathcal{H}_s$ denote the task-specific distillation heads of the teacher and student models, respectively. The extracted teacher feature $f_t$ is passed through all layers of $\mathcal{H}_t$ to produce its output. In contrast, the student feature $f_s$ is processed not only by the student's own head $\mathcal{H}_s$, but also injected into the teacher's head $\mathcal{H}_t$. Specifically, instead of entering from the first layer, $f_s$ is inserted at an intermediate fully connected layer $l$ and propagated through the remaining layers. This operation is defined as:
\begin{equation}\label{eq:logits2}
\hat{y}_{t|s} = \mathcal{H}_t^{(l:)}(f_s).
\end{equation}
Here, $\mathcal{H}_t^{(l:)}$ represents the teacher’s task-specific distillation head from layer $l$ onward. This design ensures that the student’s features are not directly forced to mimic the teacher’s low-level representations. Rather, they are aligned with higher-level decision-making processes that incorporate semantic information. To encourage the student's extracted feature $f_s$ to produce predictions similar to the teacher’s output, we apply the Kullback-Leibler divergence (KL) loss between the outputs $\hat{y}_t$ and $\hat{y}_{t|s}$. This is referred to as the distillation loss and is defined as:
\begin{equation}\label{eq:loss_kd}
\mathcal{L}_{kd} = KL (\hat{y}_t||\hat{y}_{t|s}) = \sum \hat{y}_t \log {\hat{y}_t\over \hat{y}_{t|s}}.
\end{equation}
This objective promotes the convergence of the student model toward the teacher’s semantic representations, even across heterogeneous modalities. Although $\mathcal{L}_{kd}$ facilitates this semantic alignment, additional mechanisms are required to adapt the model to the specific task. To ensure that the model learns task-specific information, we apply appropriate objective functions for regression and classification tasks, respectively. For the discrete emotion classification (DEC) task, we adopt the standard cross-entropy (CE) loss between the student’s predictions $\hat{y}_s$ and the ground truth $y$. This explicit supervision helps the student model follow the teacher’s guidance toward the correct label space, particularly under noisy or uncertain conditions. The CE loss is given by:
\begin{equation}\label{eq:loss_ce}
\mathcal{L}_{task} = CE(\hat{y}_s, y) = - \sum y \log{\hat{y}_s}.
\end{equation}
In contrast, the continuous emotion regression (CER) task is handled using the concordance correlation coefficient (CCC) loss~\cite{v2e} to measure the agreement between the predicted continuous values $\hat{y}_s$ and the ground truth $y$. This loss encourages the model to match both the mean and the variance of the predictions to the target distribution. The CCC loss is defined as:
\begin{equation}\label{eq:loss_ccc}
\mathcal{L}_{task} = CCC(\hat{y}_s, y) = {2\sigma_{{\hat{y}_s}y} \over \sigma^2_{\hat{y}_s} + \sigma^2_{y} + (\mu_{\hat{y}_s}-\mu_y)^2 },
\end{equation}
where $\sigma_{{\hat{y}_s}y}$ denotes the covariance, $\sigma_{\hat{y}_s}$ and $\sigma_y$ the variances, and $\mu_{\hat{y}_s}$ and $\mu_y$ the means.
Finally, our total loss is defined as a weighted sum of the loss terms, with $\lambda_1$, $\lambda_2$, $\lambda_3$, and $\lambda_4$ as hyperparameters:
\begin{equation}\label{eq:loss_total}
\mathcal{L} = \lambda_1 \mathcal{L}_{sim} + \lambda_2 \mathcal{L}_{unc} + \lambda_3 \mathcal{L}_{kd} + \lambda_4 \mathcal{L}_{task}.
\end{equation}
An overview of the full procedure and model architecture is illustrated in Fig.~\ref{fig1}, and the detailed training process is further summarized in Algorithm~\ref{alg:algo} as pseudo-code.

\begin{algorithm}[t]
\caption{Uncertainty-aware cross-modal knowledge distillation with prototype learning}
\label{alg:student_kd}
\begin{algorithmic}[1]

\Require \begin{minipage}[t]{\linewidth}
EEG modality samples $x_s$; \\
Visual modality samples $x_t$, Ground truth labels $y$; \\
Teacher model $\mathcal{V}$, Student model $\mathcal{E}$; \\
Loss weights $\alpha_1$, $\alpha_2$, $\alpha_3$, $\alpha_4$
\end{minipage}

\Ensure Trained student model
\State Split the dataset $\{x_s^i, x_t^i, y^i\}_{i=1}^N$ into $K$ folds;
\State Initialize teacher model $\mathcal{V}$ with pre-trained weights;
\For{$k=1, \dots, K$}
    \State \Comment{Cross-modal alignment}
    \State Extract EEG and visual features $f_s$ and $f_t$ by Eq.~\ref{eq:feat}
    \State Compute similarity loss $\mathcal{L}_{\text{sim}}(\mathbf{e}_s, \mathbf{e}_t)$ by Eq.~\ref{eq:loss_sim}
    \State Compute uncertainty loss $\mathcal{L}_{\text{unc}}(\mathbf{e}_s, \mathbf{e}_t)$ by Eq.~\ref{eq:loss_unc}
    
    \State \Comment{Cross-modal KD}
    \State Process features through respective heads and collect model predictions $\hat{y}_t$ and $\hat{y}_s$ by Eq.~\ref{eq:logits}
    \State Process student features through the teacher head and collect cross-modal prediction $\hat{y}_{t|s}$ by Eq.~\ref{eq:logits2} 
    \State Compute KD loss $\mathcal{L}_{\text{kd}}(\hat{y}_{t|s}, \hat{y}_t)$ by Eq.~\ref{eq:loss_kd}
    \State \Comment{Final loss and backpropagation}
    \State Compute task-specific loss $\mathcal{L}_{\text{task}}(\hat{y}_s, y)$ by Eq.~\ref{eq:loss_ce}, ~\ref{eq:loss_ccc}
    \State Compute total loss $\mathcal{L}$ by Eq.~\ref{eq:loss_total}
    \State Compute gradients of $\mathcal{L}$
    \State Update student model parameters 
\EndFor

\end{algorithmic}
\label{alg:algo}
\end{algorithm}


\newcolumntype{s}{>{\centering\arraybackslash}p{1.4cm}}
\newcolumntype{S}{>{\centering\arraybackslash}p{1.6cm}}
\newcolumntype{L}{>{\centering\arraybackslash}p{2.3cm}}
\begin{table*}[!t]
    \centering
    \caption{Performance comparison of our proposed method with baseline unimodal and multimodal methods for discrete emotion classification (DEC) and continuous emotion regression (CER) tasks on the MAHNOB-HCI dataset~\cite{mahnob}.}
    \label{table:overall_1}
    \renewcommand{\arraystretch}{1.1}  
    \begin{threeparttable}
    \begin{tabular}{c L  s s s s  S S S}
        \toprule[1.3pt]
        
        \multicolumn{2}{c}{%
            \multirow{3}{*}[-0.5em]{\centering Method}%
        }%
        & \multicolumn{4}{c}{DEC}
        & \multicolumn{3}{c}{CER} \\
        \cmidrule(lr){3-6} \cmidrule(lr){7-9}

        & &
        \multicolumn{2}{c}{Arousal} & \multicolumn{2}{c}{Valence}
        & \multicolumn{3}{c}{Valence} \\
        \cmidrule(lr){3-4} \cmidrule(lr){5-6} \cmidrule(lr){7-9}

        & &
        Acc. $\pm$ std. & F1. $\pm$ std. & Acc. $\pm$ std. & F1. $\pm$ std.
        & RMSE $\pm$ std. & PCC $\pm$ std. & CCC $\pm$ std. \\
        \midrule[1.1pt]

        \multirow{4}{*}{\centering Unimodal}
         & DeepConvNet\cite{deepconvnet}    & 44.0$\pm$1.52 & 42.1$\pm$1.77 & 45.7$\pm$2.36 & 44.2$\pm$1.76 & 0.049$\pm$0.030 & 0.323$\pm$0.045 & 0.255$\pm$0.034
         \\
         & EEGNet\cite{eegnet}              & 43.6$\pm$1.87 & 47.4$\pm$1.46 & 43.1$\pm$2.01 & 49.2$\pm$2.94 & 0.047$\pm$0.035 & 0.316$\pm$0.078 & 0.301$\pm$0.079
         \\
         & EEGFormer\cite{eegformer}     & 45.2$\pm$3.49 & 40.8$\pm$3.24 & 48.1$\pm$5.08 & 50.2$\pm$4.82 & 0.050$\pm$0.029 & 0.334$\pm$0.056 & 0.238$\pm$0.072
         \\
         & MASA-TCN~\cite{masa}             & 46.4$\pm$1.32 & 48.7$\pm$1.56 & 45.8$\pm$2.67 & 50.9$\pm$4.65 & 0.058$\pm$0.035 & 0.390$\pm$0.044 & 0.338$\pm$0.045
         \\
        \midrule

        \multirow{4}{*}{\centering Multimodal}
         & CAFNet~\cite{caf}                & \textbf{59.4$\pm$1.17} & 59.6$\pm$1.83 & 55.6$\pm$3.02 & 57.2$\pm$2.59 & 0.046$\pm$0.062 & 0.368$\pm$0.058 & 0.309$\pm$0.060
         \\
         & Visual-to-EEG~\cite{v2e}         & 53.9$\pm$3.34 & 54.6$\pm$2.71 & 50.3$\pm$1.42 & 54.6$\pm$1.12 & 0.055$\pm$0.028 & \textbf{0.462$\pm$0.052} & \textbf{0.372$\pm$0.084}
         \\
         & EmotionKD~\cite{ekd}             & 56.8$\pm$2.57 & 52.2$\pm$3.14 & 49.9$\pm$2.24 & 52.0$\pm$2.28 & 0.052$\pm$0.034 & 0.422$\pm$0.085 & 0.310$\pm$0.056
         \\
         & \textbf{Ours}                    & 57.1$\pm$2.38 & \textbf{60.0$\pm$1.45} & \textbf{57.9$\pm$3.21} & \textbf{59.4$\pm$2.47}  & \textbf{0.043$\pm$0.027} & 0.449$\pm$0.075 & 0.359$\pm$0.069
         \\
        \bottomrule[1.3pt]
    \end{tabular}
    \begin{tablenotes}
        \item Acc.: accuracy ($\%$), F1.: F1 score ($\%$), RMSE: root mean squared error, PCC: Pearson correlation coefficient, CCC: concordance correlation coefficient
    \end{tablenotes}
    \end{threeparttable}
\end{table*}

\section{EXPERIMENTS}

\subsection{Dataset}
MAHNOB-HCI~\cite{mahnob} is a widely used multimodal dataset for emotion recognition and implicit tagging research based on behavioral and physiological responses. It includes recordings from 30 participants who watched affective video clips lasting 34 to 117 seconds. During each session, EEG signals were recorded using 32 electrodes, and behavior was captured with 6 cameras. Subjects 12, 15, and 26 did not complete data collection, so only 27 participants are used in this study. Each trial includes self-reported ratings from 1 to 9 for arousal, valence, dominance, and predictability, and valence was additionally annotated by experts to capture temporal emotional dynamics. Therefore, the dataset enables research on both emotion classification and regression tasks.

\subsection{Task}
Emotion recognition tasks are generally categorized into two main approaches: discrete emotion classification and continuous emotion regression~\cite{aBCI}. DEC involves categorizing emotional states into predefined classes such as happiness, sadness, anger, or fear. In contrast, CER represents emotions as points in a continuous space, typically defined by the valence–arousal model. While DEC is often easier to interpret and evaluate in terms of categorical emotion labels, CER enables the modeling of subtle and dynamic emotional changes for more precise emotional analysis.

\subsection{Implementation Details}
The overall architecture employs an EEG-Conformer~\cite{eegconformer} and a CNN-based transformer as the feature extractors for the student and teacher models, respectively.
The model was trained using the Adam optimizer with the learning rate ranging from $10^{-3}$ to $10^{-6}$ for 100 epochs. The dataset was partitioned using 5-fold cross-validation with a group-by-trial strategy. 
To improve generalization, early stopping with a patience of 20 epochs was applied during training in each fold. The entire framework was implemented in PyTorch and trained on NVIDIA GPUs to accelerate computation.


\section{RESULTS}
In this section, we first compare our method with baseline approaches on both discrete and continuous emotion recognition tasks. For the DEC task, we evaluate performance using accuracy and macro F1 score, while for the CER task, we report root mean squared error (RMSE), Pearson correlation coefficient (PCC), and concordance correlation coefficient (CCC) as evaluation metrics. We then present feature space visualizations to demonstrate the effectiveness of our approach, followed by an ablation study analyzing the contribution of each loss component of our framework.

\subsection{Classification Performance}
To validate the effectiveness of our proposed method, we compare it with both unimodal and multimodal baselines on the MAHNOB-HCI dataset. Table~\ref{table:overall_1} summarizes the classification performance for the DEC task and the regression performance for the CER task. Here, we report the DEC results, focusing on accuracy and F1 scores for arousal and valence, where our method shows consistent improvements over unimodal baselines. We compare against DeepConvNet~\cite{deepconvnet}, EEGNet~\cite{eegnet}, EEGFormer~\cite{eegformer}, and MASA-TCN~\cite{masa}, including models designed for EEG-based emotion recognition and those commonly used in general EEG modeling, among which MASA-TCN demonstrates competitive performance. However, our model achieves superior results, with 57.1$\%$ accuracy and the highest F1 score of 60$\%$ in arousal classification.
Beyond unimodal models, our method continues to demonstrate strong performance compared to multimodal baselines, such as CAFNet~\cite{caf}, Visual-to-EEG~\cite{v2e}, and EmotionKD~\cite{ekd}, which incorporate additional modalities for affective state decoding from EEG signals.
Although CAFNet reports higher accuracy in arousal classification, our model records the highest F1 score for arousal and outperforms all baselines in both accuracy and F1 score for valence classification. This comprehensive superiority indicates that our method achieves the most balanced and reliable performance across emotion categories.

\begin{figure}[!t]
\centerline{\includegraphics[width=\columnwidth]{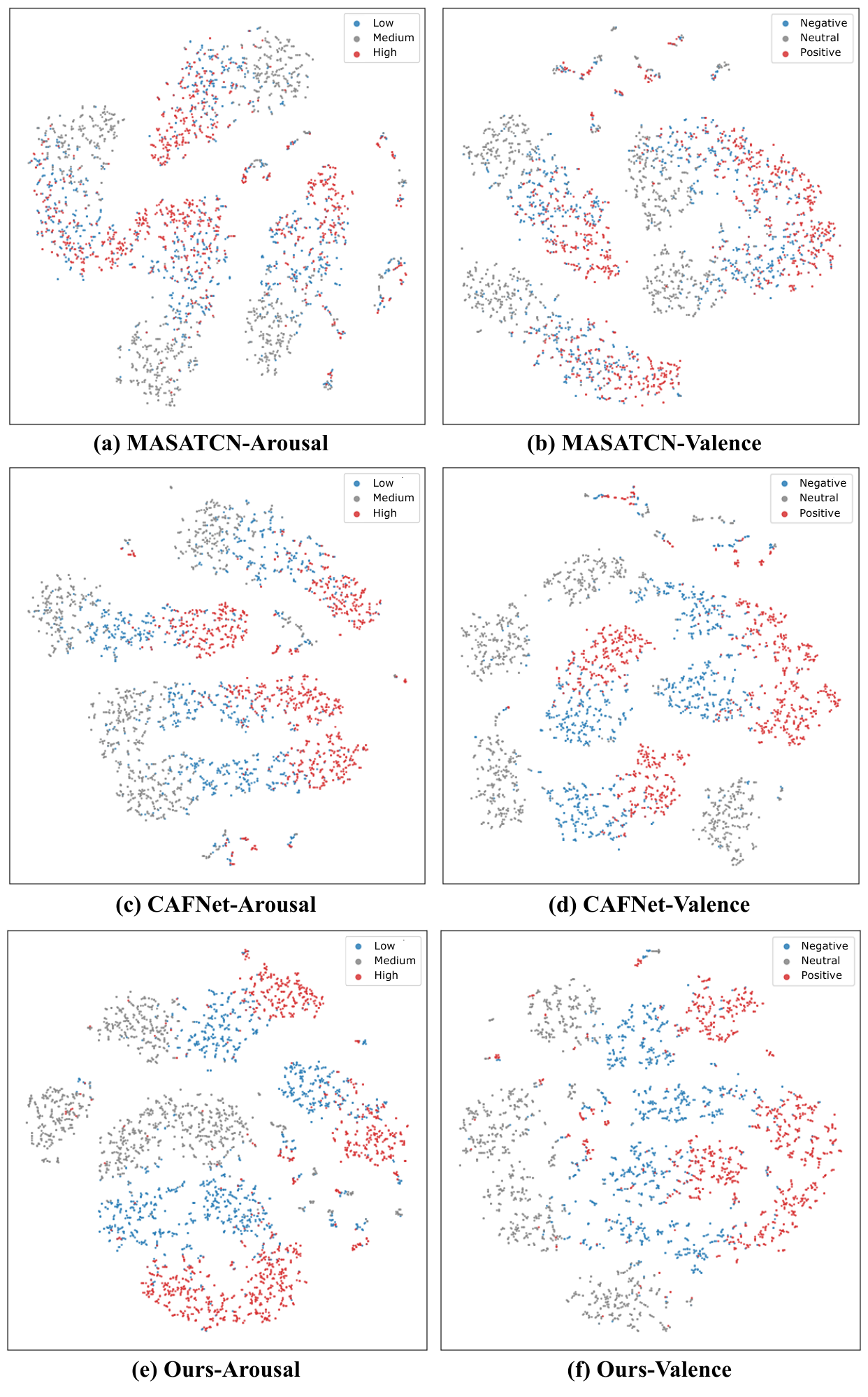}}
\caption{Visualization of the learned feature distribution. For comparison, we present the features learned by the unimodal sota model MASA-TCN (top), the multimodal sota CAFNet (middle), and our proposed cross-modal KD method (bottom) for both the arousal and valence classification tasks.}
\label{fig:tsne_comparison}
\end{figure}

\subsection{Regression Performance}
We evaluate CER performance using RMSE, PCC, and CCC metrics for valence prediction, as summarized in Table~\ref{table:overall_1} based on the MAHNOB-HCI dataset. Lower RMSE indicates better prediction accuracy, while higher PCC reflects stronger correlation with the ground truth. Among all compared models, our proposed method achieves the lowest RMSE of 0.043, and the second-highest PCC and CCC scores of 0.449 and 0.359, respectively.
Compared to unimodal baselines, our model shows clear improvements across all three metrics. Among unimodal methods, MASA-TCN~\cite{masa} demonstrates competitive performance but still falls short of our approach. In the multimodal setting, Visual-to-EEG~\cite{v2e} demonstrates strong correlation metrics, although its performance is less stable in terms of prediction error. In contrast, our model attains superior balance, achieving both the lowest RMSE and strong correlation metrics.
Taken together, the results demonstrate that our approach effectively captures both discrete and continuous emotional representations. We attribute this gain to extracting semantically aligned features and making robust predictions under noisy EEG conditions. 

\newcolumntype{S}{>{\centering\arraybackslash}p{0.5cm}}
\newcolumntype{C}{>{\centering\arraybackslash}p{1.35cm}}
\begin{table}[t]
    \centering
    \caption{Ablation study of the proposed loss components on the overall performance for the DEC task.}
    \label{table:ablation}
    \renewcommand{\arraystretch}{1}
    \begin{threeparttable}
    \begin{tabular}{S@{\hspace{10pt}}S@{\hspace{10pt}}S C@{\hspace{3pt}} C@{\hspace{4pt}} C@{\hspace{3pt}} C} 
        \toprule[1.3pt]
        \multirow{2}{*}{$\mathcal{L}_{sim}$} & 
        \multirow{2}{*}{$\mathcal{L}_{unc}$} & 
        \multirow{2}{*}{$\mathcal{L}_{kd}$} & 
        \multicolumn{2}{c}{Arousal} & 
        \multicolumn{2}{c}{Valence} \\
        \cmidrule(lr){4-5} \cmidrule(lr){6-7}
        & & & Acc. $\pm$ std. & F1. $\pm$ std. & Acc. $\pm$ std. & F1. $\pm$ std. \\ 
        \midrule[1.1pt]
        \checkmark &  &                       & 42.8$\pm$4.94 & 44.2$\pm$7.45 & 41.6$\pm$4.36 & 41.9$\pm$4.73 \\
        & \checkmark &                        & 43.0$\pm$1.80 & 43.2$\pm$2.12 & 42.8$\pm$2.97 & 42.9$\pm$3.32 \\
        &  & \checkmark                       & 45.3$\pm$2.63 & 47.3$\pm$2.88 & 44.4$\pm$7.91 & 43.6$\pm$2.11 \\
        \checkmark & \checkmark &             & 51.0$\pm$3.61 & 53.7$\pm$0.94 & 43.9$\pm$6.95 & 44.1$\pm$7.24 \\
        \checkmark &  & \checkmark            & 54.6$\pm$3.97 & 53.2$\pm$2.26 & 48.4$\pm$4.51 & 48.5$\pm$4.66 \\
        & \checkmark & \checkmark             & \textbf{58.6}$\pm$1.79 & 58.9$\pm$1.28 & 51.4$\pm$8.23 & 57.3$\pm$2.79 \\
        \checkmark & \checkmark & \checkmark  & 57.1$\pm$2.38 & \textbf{60.0$\pm$1.45} & \textbf{57.9$\pm$3.21} & \textbf{59.4$\pm$2.47} \\
        \bottomrule[1.3pt]       
    \end{tabular}
    \end{threeparttable}
\end{table}

\newcolumntype{S}{>{\centering\arraybackslash}p{0.5cm}} 
\newcolumntype{W}{>{\centering\arraybackslash}p{1.6cm}}
\begin{table}[!t]
    \centering
    \caption{Ablation study of the proposed loss components on the overall performance for the CER task.}
    \label{table:ablation2}
    \renewcommand{\arraystretch}{1}
    \begin{threeparttable}
    \begin{tabular}{S S S W W W}  
        \toprule[1.3pt]
        \multirow{2}{*}{$\mathcal{L}_{sim}$} & \multirow{2}{*}{$\mathcal{L}_{unc}$} & \multirow{2}{*}{$\mathcal{L}_{kd}$} & \multicolumn{3}{c}{Valence} \\
        \cmidrule(lr){4-6}
        & & & RMSE $\pm$ std. & PCC $\pm$ std. & CCC $\pm$ std. \\ 
        \midrule[1.1pt]
        \checkmark &  &                          & 0.057$\pm$0.018 & 0.251$\pm$0.064 & 0.226$\pm$0.052 \\
        & \checkmark &                           & 0.046$\pm$0.012 & 0.300$\pm$0.068 & 0.266$\pm$0.082  \\
        &  & \checkmark                          & 0.053$\pm$0.016 & 0.267$\pm$0.037 & 0.243$\pm$0.077 \\
        \checkmark & \checkmark &                & 0.052$\pm$0.020 & 0.315$\pm$0.057 & 0.305$\pm$0.058 \\
        \checkmark &  & \checkmark               & 0.050$\pm$0.012 & 0.357$\pm$0.066 & 0.343$\pm$0.072\\
        & \checkmark & \checkmark                & 0.049$\pm$0.017 & 0.406$\pm$0.080  & \textbf{0.369$\pm$0.078}\\
        \checkmark & \checkmark & \checkmark     & \textbf{0.043$\pm$0.027} & \textbf{0.449$\pm$0.075} & 0.359$\pm$0.069  \\
        \bottomrule[1.3pt]      
    \end{tabular}
    \end{threeparttable}
\end{table}

\subsection{Feature Space Visualization}
In order to verify the effectiveness of our framework, we visualize the learned feature distributions projected onto a two-dimensional space, as shown in Fig.~\ref{fig:tsne_comparison}.
We compare our method against a strong unimodal baseline and a competitive multimodal model, focusing on feature representations learned during arousal and valence classification tasks.
We first examine the projected feature spaces learned by MASA-TCN in Fig. 2(a) and 2(b), which reveal scattered and overlapping clusters. This indicates weak separability between low and high emotional states, suggesting limited discriminative power in the learned EEG representations. Fig. 2(c) and 2(d) show the feature spaces learned by CAFNet, demonstrating improved clustering compared to MASA-TCN, particularly for arousal. However, class overlap still persists highlighting challenges in modeling subtle emotional variations. In contrast, Fig. 2(e) and 2(f) illustrate that our method yields well-separated and compact clusters in both emotional dimensions. In particular, the valence feature distributions demonstrate clear separation among negative, neutral, and positive emotional states.  
Moreover, they exhibit tight class grouping, which reflects stronger consistency and enhanced discriminative capability. These visual patterns support the quantitative results reported in Table~\ref{table:overall_1}, highlighting that our method consistently outperforms MASA-TCN and CAFNet across most evaluation metrics.

\subsection{Ablation Study}
To examine the contribution of each loss component, we conducted ablation studies, as summarized in Table~\ref{table:ablation} (DEC) and Table~\ref{table:ablation2} (CER). We evaluated three loss terms: similarity-based loss $\mathcal{L}_{sim}$, uncertainty-aware loss $\mathcal{L}_{unc}$, and knowledge distillation loss $\mathcal{L}_{kd}$.
$\mathcal{L}_{kd}$ shows the strongest individual and combined effect across both DEC and CER tasks, highlighting its role in effective teacher supervision.
Meanwhile, $\mathcal{L}_{unc}$ improves robustness to ambiguous EEG signals and shows strong synergy with $\mathcal{L}_{kd}$, with their combination yielding the highest arousal accuracy. However, the best overall performance is achieved when all three components are used together.
In the CER task, $\mathcal{L}_{unc}$ exhibits the most significant impact applied in isolation.
Among the two-loss combinations, the pair of $\mathcal{L}_{sim}$ and $\mathcal{L}_{kd}$ delivers the strongest performance, indicating effective interaction between feature similarity and teacher supervision. Still, this configuration falls short compared to the full combination of all three losses.
Across both DEC and CER tasks, $\mathcal{L}_{kd}$ consistently proves to be the most influential component. This trend underscores its central role in enabling effective cross-modal supervision. Incorporating all three objectives leads to more robust and generalizable emotion recognition, especially under ambiguous EEG conditions.

\section{CONCLUSION}
In this paper, we propose a cross-modal knowledge distillation framework that leverages the visual modality to guide EEG-based emotion recognition by mitigating the noise and ambiguity of EEG signals through representation alignment, uncertainty estimation, and semantic consistency.
Extensive experiments on the MAHNOB-HCI dataset provide a comparison between our framework and existing unimodal and multimodal baselines. While CAFNet and Visual-to-EEG show strengths in specific DEC and CER metrics respectively, their performance lacks consistency across tasks. 
In contrast, our approach achieves the best results across most metrics, demonstrating stronger overall robustness, as confirmed by feature space visualizations. 
In the ablation study, although $\mathcal{L}_{sim}$ provides a basic alignment signal, its limited effect when used alone suggests that effective representation learning requires not only feature similarity but also semantic guidance.
In future work, we plan to explore alignment strategies that account for the structure and reliability of each modality to further enhance generalization.
Overall, our findings underscore the value of semantically guided objectives for effective knowledge transfer, promoting robust and interpretable EEG-based BCI systems.

\addtolength{\textheight}{-12cm}   


\bibliographystyle{IEEEtran}
\bibliography{reference}

\end{document}